\documentclass[review]{elsarticle}

\usepackage{hyperref}

\journal{Journal of \LaTeX\ Templates}

\usepackage{hyperref}
\usepackage{amsmath}
\usepackage{float}
\usepackage{amsmath,amssymb,amsfonts}
\usepackage{algorithm,algorithmic}
\usepackage{algorithmic}
\usepackage{graphicx}
\usepackage{textcomp}
\usepackage{xcolor}
\usepackage{adjustbox}
\newcommand{\RNum}[1]{\uppercase\expandafter{\romannumeral #1\relax}}

\begin{document}

\begin{frontmatter}

\title{Generalised agent for solving higher board states of tic tac toe using Reinforcement Learning}

\author{Bhavuk Kalra}
\address{Department of Computer Engineering, Thapar Institute of Engineering and Technology, Patiala 147001, India, bkalra\_be18@thapar.edu}




\begin{abstract}
Tic Tac Toe is amongst the most well-known games. It has already been shown that it is a biased game, giving more chances to win for the first player leaving only a draw or a loss as possibilities for the opponent, assuming both the players play optimally. Thus on average majority of the games played result in a draw. The majority of the latest research on how to solve a tic tac toe board state employs strategies such as Genetic Algorithms, Neural Networks, Co-Evolution, and Evolutionary Programming. But these approaches deal with a trivial board state of 3X3 and very little research has been done for a generalized algorithm to solve 4X4,5X5,6X6 and many higher states. Even though an algorithm exists which is Min-Max but it takes a lot of time in coming up with an ideal move due to its recursive nature of implementation. A Sample has been created on this link \url{https://bk-tic-tac-toe.herokuapp.com/} to prove this fact.
This is the main problem that this study is aimed at solving i.e providing a generalized algorithm(Approximate method, Learning-Based) for higher board states of tic tac toe to make precise moves in a short period. Also, the code changes needed to accommodate higher board states will be nominal. The idea is to pose the tic tac toe game as a well-posed learning problem. The study and its results are promising, giving a high win to draw ratio with each epoch of training. This study could also be encouraging for other researchers to apply the same algorithm to other similar board games like Minesweeper, Chess, and GO for finding efficient strategies and comparing the results.
\end{abstract}

\begin{keyword}
\texttt Reinforcement Learning, Game Theory,
\MSC[2010] 00-01\sep  99-00
\end{keyword}

\end{frontmatter}

\section{Introduction}

\begin{figure}[h]
    \centering
    \includegraphics[width = 7cm]{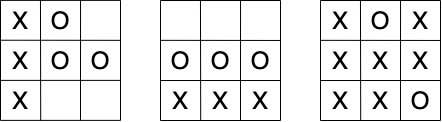}
    \caption{Examples of invalid board-states}
    \label{InvalidBoards}
\end{figure}

In the research done in \cite{games_possible} it is stated that in tic tac toe 362,880 different possibilities can be solved using a searching algorithm on a 3x3 grid counting invalid games and games where the game should have already ended from a win. Some of the invalid board states are shown in Figure \ref{InvalidBoards}. There are 255,168 possible valid games in total(excluding symmetry).

\begin{figure}[h]
    \centering
    \includegraphics[width = 4.5cm]{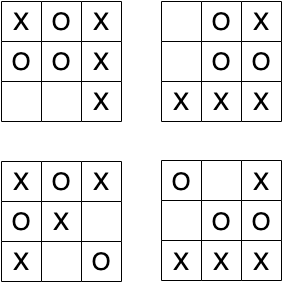}
    \caption{Examples of valid terminal-states(won by X)}
    \label{ValidBoards}
\end{figure}

Continuing on the research about the number of boards states that are possible, in the research done in \cite{chou2013using} it has been claimed that the first player wins 131184 of the permutations, the second player wins 77904, and the residual 46080 results in a draw. 958 unique terminal configurations can be reached when a winner is found assuming X is the first player. Figure \ref{ValidBoards} lists out a few possibilities out of these 958 games. 

\subsection{Pre-existing algorithmic solutions}
It is to be noted that tic tac toe is a biased game towards the first player and gives more winning chances to the player who starts the game leaving the opponent only a loss or draw as possibilities. However, it is already known and specified in \cite{singh2014never}  that if both players play rationally, the game would end in a draw. A player employs a strategy for playing the game in such a way that he either wins or draws but never loses. Efforts have been made on creating a fast and effective tic tac toe playing algorithm and similar other board games which employs a state-space search, ex. A modification of the Tic Tac presented in \cite{maureT}, Below are listed some of the major contributions to the creation of various techniques for the game of tic tac toe. \cite{sutton2018reinforcement}  \par

A major contribution made recently in this field is of Google Deepmind's alphaGO which had a huge impact in this field by introducing a generalized deep reinforcement learning to the world of strategy board games by creating agents that could play many different Atari and similar board games like tic tac toe(chess, shogi and GO) and even managed to outplay the best chess engine at that time(StockFish 8) as stated in  \cite{mnih2015human} and \cite{silver2017mastering}.But the actual algorithm behind the agent has not been made fully public yet, which supposedly only required training of only 4 hours on google's 64 GPU and 19 CPU using indirect feedback. Presumably, only the rules of the game were given as input the rest(strategies, tactics), the algorithm learned itself in the form of Self-Play. Similar was the adoption using qlearning utilized in this citation \cite{qlearning}. A similar study was covered in \cite{wang2021searching}, where an attempt was made to improve the convergence rate. On similar grounds to Reinforcement Learning an interesting approach was taken to impose the human-like aspect to it and named it theory-based reinforcement learning. \cite{tsividis2021human} \cite{wang2021adaptive}. AlphaGO like approach but implemented via one-hot encoding based vectors\cite{gu2021enhanced} \cite{scheiermann2022alphazero} \par 

In \cite{bhatt2007002evolution} the authors did an interesting study on evolving several strategies with ”No-Loss” for the game of tic tac toe and found 72,657 solutions by their implementation of a Genetic Algorithm. This went above what we already knew that there are existing No-Loss strategies but this highlighted upon what estimated number of them could exist.\par

To find the best pass for a 3X3 tic tac toe game researchers in \cite{bits_pilani} study employed hamming distance classifier based neural network, with time complexity for the algorithm being $O(n^3)$ where n is the number of cells(3X3, n = 9). Which would take a lot of time if the tic tac toe game were to be scaled up to 4X4,5X5 etc.A similar Methodology is employed by researchers in the paper using evolutionary programming in \cite{fogel1993using}. \par

A more recent study explored a more interactive approach on how the game of tic tac toe should be played using drones. In that research the authors explored the algorithms such as QL algorithm, SV and SARSA. \cite{karmanova2021swarmplay}

One of the more common solutions to the tic tac toe problem is using the Min-Max search algorithm. \cite{doi:10.1080/08839514.2021.1934265}

Borovska in his paper \cite{acm_borovska} has discussed the reliability of the Min-Max algorithm. This paper also discusses an Optimization approach($\alpha-\beta$ pruning) and the advantages and disadvantages of the Min-Max algorithm, which works on a basis of trying to minimize the loss and maximize the gain at each step of the way down the search tree in a recursive fashion, by attributing certain characteristics for the game. An optimization to it is covered in \cite{keswani2022convergent}, \cite{cai2022accelerated} and \cite{abernethy2021last}.\par

\begin{figure}[h]
    \centering
    \includegraphics[width = 8.8cm]{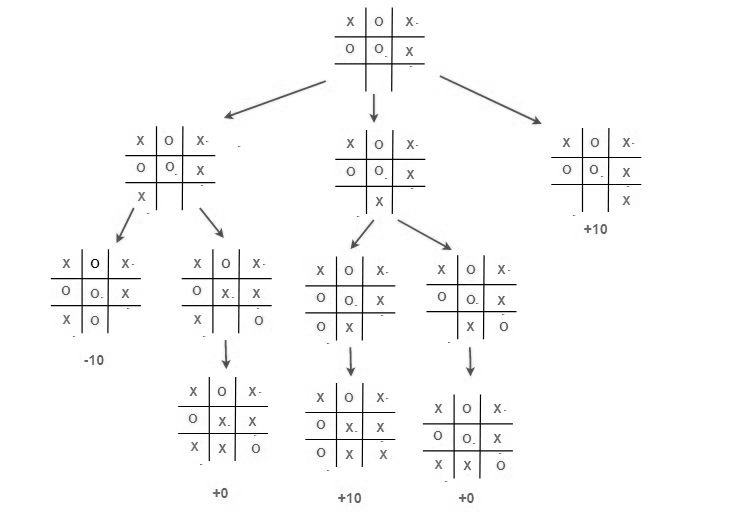}
    \caption{Search tree for a specific game state using +10, 0 and -10 as terminal utility values}
    \label{searchTree}
\end{figure}

The researchers in \cite{5205101} paper discussed a decision tree-based approach for implementing no loss state in tic tac toe.
A Theoretical error is also highlighted in this paper that, when both players play correctly the Min-Max does have an ideal No-Loss strategy However, when the opponent plays non-optimally Min-Max has been shown to play non-optimal moves, dragging the game out for one extra move resulting to succeed at the next state rather than the planned state. \par

The Min-Max algorithm explores the game tree in detail first so with increasing complexity for higher board sizes the game tree in Min-Max becomes very large and it takes a long time to come up with a solution, especially at the starting when the whole board is empty. The time complexity of the Min-Max algorithm is $O(b^{m})$ and the space complexity is $O(bm)$ where $m$ is the maximum depth of the tree(9 for tic tac toe) and $b$ is the number of legal moves at each node. In the next section, we will discuss how we can try to reduce this time for board sizes of 3X3, 4X4, and so on for the tic tac toe game specifically. Though the algorithm can be implemented for other board games as well and scaled up for higher board sizes with minimalist code changes. This reduction in time for different board sizes can be achieved by defining tic tac toe as a \emph{well-posed problem} to improve decision taking time for an agent. \par

\section{Well-Posed learning problems}

"A computer program is said to learn from experience E with respect to some class of tasks T and performance measure P, if its performance at tasks in T, as measured by P, improves with experience E"\cite{TomMMitchell}. \par

Through experience gained from playing games against itself, a computer algorithm that learns to play will boost its competitiveness as calculated by its chances of winning a set of tic tac toe games.

The same terminology is used in this paper as used in \cite{TomMMitchell} and according to it, 

There are three main features of a learning system: 
\begin{enumerate}
  \item The classification of task, denoted by T
  \item The efficiency metric that needs to be improved, denoted by P
  \item The root of knowledge and source of experience, denoted by E
\end{enumerate}

\section{Designing a Tic-Tac-Toe learning system}

\subsection{Choosing the training experience }

One significant feature of a training experience is whether it offers direct or indirect guidance on the system's decisions. As direct training examples, individual tic tac toe board states and the accurate action for each are used. Conversely, the agent might have access to only indirect knowledge such as step sequences and game outcomes from previous games. In this case, the learner is confronted with the issue of \emph{credit assignment}. That is determining how much credit or blame each step in the series deserves for the final result. Credit allocation is an especially difficult problem because even though early moves are optimal, the game can be lost if they are followed by bad moves later on. As a result, learning through direct input as training is normally preferable to learning from an indirect one, where only a series of steps and the final game result is known since it includes the board states and best move for each of them right at the start and the agent does not have to figure out the strategy behind a particular move being played from scratch as it has to in indirect feedback. Here in this paper, an approach with Indirect learning is employed i.e The machine is being trained by letting it play games against itself. This has the benefit of requiring no additional input/strategy i.e the agent will only know the rules of the game and
as a result, it enables the machine to produce as much training data as time allows. This is a very similar approach as used by Google's Deepmind AlphaGO for playing chess.\par

The extent to which the learner affects the sequence of training example set is a second vital feature of the training experience. There are many possibilities for implementing this attribute some of them are.

\begin{enumerate}
  \item Depend entirely on the input Dataset for the selection and the best move for each of the Board States
  
  \item   The board states that seem particularly ambiguous
  (those with the same Utility Value) might be proposed by the agent and then it queries the database for the correct move

  \item  The agent can fully monitor both board states and indirect training configurations when it learns to play against itself without a reference database
\end{enumerate}
\par

The way the distribution of examples is represented, which is a measure of final performance P for the system, is the third important aspect of the training experience. Learning is most effective when training examples follow a pattern that is identical to the pattern of potential future test examples. The efficiency metric P in our tic tac toe learning scenario is the percentage of games won by the system. If the training experience E consists solely of games played against itself, there is a good chance that it will not be entirely representative of the total range of situations in which it will be tested against later. The distribution of training and test samples is identical is the critical assumption that the majority of current machine learning theory is based upon. But this assumption can be problematic in some situations where the training experience of agent E does not account for it.

\subsection{Selecting a Target Function}

The performance system's next implementation choice is to determine exactly what kind of data will be collected and how it will be used. The correct move for any defined board state is determined by this method which is used to choose the type of knowledge to be learned. For evaluating the utility value of some particular board state a target function is used. This function helps to determine whether it is desirable or not for a state reached after a certain move. By comparing their utility values through a predefined system for giving values to some definitive board states (Win, Draw, Loss), we compare the desirability of two board states. \par

We call this target function $V$ denoted by $V:B \xrightarrow[]{}\mathbb{R}$, Here $V$ is a many-one mapping between any board state(legal) to some real number value(denoted by set $\mathbb{R}$)

This target function $V$ must be defined in such a way that it gives higher values to desirable board states. If we can make the system learn such a target function then the system just has to generate a set of possible board states(legal) from the current state(using a $moveGenerator$ function), compare their utility value, and choose the best out of them, and therefore choosing the best legal move. \par

For a random board state $b$ in $B$ the target value $V(b)$ can be stated as the following

\[V(b)  = \begin{cases} 
          100 & \text{if final board state b is won} \\
          -100 & \text{if final board state b is lost}\\
          0 & \text{if final board state b is drawn}\\
          V(b') & \text{if b is a not a final board state}\\
       \end{cases}
    \]
    
Where $b'$ is the optimal final board state
of the board that can be obtained from $b$(an intermediate state) after optimum play to the end of the session.

This is a recursive definition of the target function $V(b)$. But this definition is not easily computable by our agent(especially for starting and intermediate states) as it requires traversing the game tree from the current node b to all possible configurations. Hence it is not a viable option and some optimization must be done to get the utility values for intermediate states without exploring the whole search tree. In other words, it is a non-operational definition. In this scenario, The main goal of learning is to create a strategic functional overview of $V$ that our learning software will use to evaluate states and choose appropriate behavior under reasonable time limits. In this sense, this has reduced the challenge of learning to the finding of an operational definition of the optimal target function $V$. \par

Here to achieve some approximation to the ideal target function $V$ we will use an approximation function. This function will be distinguished from the ideal target function $V$ and will be implemented in a separate class to maintain a low coupling architecture for the final design. The symbol $\hat{V}$ is used to correspond to the approximation function that the agent will use.

\subsection{Selection of Target Function Representation}

After the ideal target function, $V$ is defined, the approximation function $\hat{V}$ representation used by our agent will be defined in this section. The following \emph{board features} are considered for a specific intermediate board state. A linear combination of these features will account for the definition of the function $\hat{V}$:

\begin{itemize}

  \item  $x_1$: Number of Instances of Player-1's Symbol(X) within an open row and column. Examples in Figure \ref{x_in_open_row} and \ref{x_in_open_col}
  
  \begin{figure}[H]
    \centering
    \includegraphics[width = 4cm]{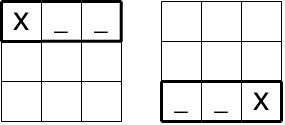}
    \caption{Some examples of X within an open row}
    \label{x_in_open_row}
\end{figure}

\begin{figure}[h]
    \centering
    \includegraphics[width = 4.5cm]{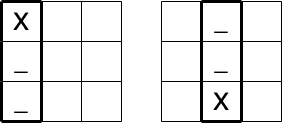}
    \caption{Some examples of X within an open column}
    \label{x_in_open_col}
\end{figure}
  
  \item $x_2$: No. of Instances of Player-2's Symbol(O) in a row within an open row and column. Examples in Figure \ref{o_in_open_row} and \ref{o_in_open_col}.

    \begin{figure}[h]
    \centering
    \includegraphics[width = 4cm]{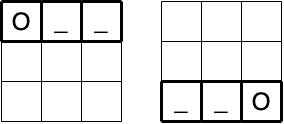}
    \caption{Some examples of O within an open row}
    \label{o_in_open_row}
\end{figure}

\begin{figure}[h]
    \centering
    \includegraphics[width = 4cm]{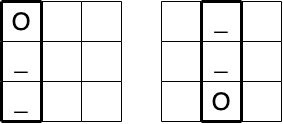}
    \caption{Some examples of O within an open column}
    \label{o_in_open_col}
\end{figure}

  \item $x_3$: No. of instances of 2 consecutive Player-1's Symbol(X) in a row. Examples in Figure \ref{2_x_in_row}
     \begin{figure}[H]
    \centering
    \includegraphics[width = 4.5cm]{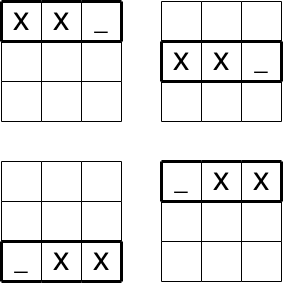}
    \caption{Some examples consecutive X in a row}
    \label{2_x_in_row}
\end{figure}

 \item $x_4$:No. of instances of 2 consecutive Player-2's Symbol(O) in a row. Examples in Figure \ref{2_o_in_row}
 
     \begin{figure}[H]
    \centering
    \includegraphics[width = 4.5cm]{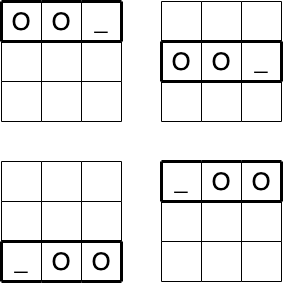}
    \caption{Some examples consecutive O in a row}
    \label{2_o_in_row}
\end{figure}

\item $x_5$:No. of instances of 3 Player-1's Symbol(X) in a row with an open box. Examples in Figure \ref{x_in_open_box}

     \begin{figure}[H]
    \centering
    \includegraphics[width = 6cm]{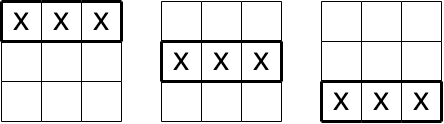}
    \caption{Possibilities of all X in a row}
    \label{x_in_open_box}
\end{figure}

\item $x_6$:No. of instances of 3 Player-2's Symbol(O) in a row with an open box. Examples in Figure \ref{o_in_open_box}

     \begin{figure}[h]
    \centering
    \includegraphics[width = 6cm]{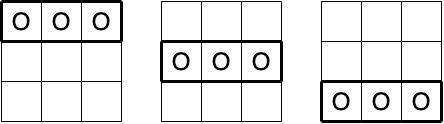}
    \caption{Possibilities of all O in a row}
    \label{o_in_open_box}

\end{figure}

\end{itemize}

With all the features that will be extracted from a board state defined, the overall \emph{feature vector} $\overrightarrow{X}$ will be($x_0 = 1$ is for biasing) as explained in \cite{goodfellow2016machine}

$$ \overrightarrow{X} = \begin{bmatrix}1\\x_1\\x_2\\x_3\\x_4\\x_5\\x_6\end{bmatrix}$$

and corresponding \emph{weight vector} will be

$$ \overrightarrow{W} = \begin{bmatrix}w_0\\w_1\\w_2\\w_3\\w_4\\w_5\\w_6\end{bmatrix}$$

As a result, The learning program will represent $\hat{V}(b) = \overrightarrow{X}.\overrightarrow{W}$ as a linear function of the form

\begin{equation}
    \hat{V}(b) = w_0 + w_1x_1 + w_2x_2+ w_3x_3+ w_4x_4 + w_5x_5 + w_6x_6
\end{equation}


In machine learning terms $w_0$ through $w_6$ are called weights.
They decide the relative significance of the feature for what they are defined for and a combined linear combination of them gives the utility of a particular board state. $w_1$ to $w_6$ are normal weights given to predefined features of the board above. While $w_0$ is a constant that adds to the importance of the board.\cite{TomMMitchell}.

\subsection{Choosing an Algorithm for Function Approximation}

A collection of training examples is required to learn the target function $\hat{V}$. Each training example will include an ordered pair of the form $(b, V_{train}(b))$ where b is the board state(represented in the form of feature vector) and $V_{train}(b)$ is the training value for b. For instance, given below is an example, where X has won, for which the target function value $V_{train}(b)$ is $+100$.

\begin{figure}[h]
    \centering
    \includegraphics[width = 2.5cm]{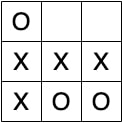}
    \caption{Example board state for feature vector calculation}
    \label{fig:searchTree}
    
\end{figure}

$$((x_1 = 0,x_2 = 1, x_3 = 0, x_4 = 0, x_5 = 1, x_6 = 0), +100)$$

Following that, we'll look at a method for extracting unique training scenarios from the learner's indirect training experience and then changing the weights $w_i$ to best suit these training scenarios

\subsection{Estimating intermediate training values}

It's relatively simple to assign utility values to board states that have definite outcomes (Loss, Won, Drawn), but it's less clear how to manage the intermediate board states that occur in a game. It's also worth noting that just because the game ends in a win or a loss, it doesn't mean all of the board states along the way were good or bad. To put it another way, a game ending in a victory does not imply that the winning player played perfectly, and a game-ending in a loss does not imply that the opponent made no good moves.

The approach that is used in this paper for estimating the intermediate state's training values is to assign it to the training value of the successor states. In more technical terms if $Successor(b)$ denotes the next board state following $b$ and the agent's current approximation to $V$ is $\hat{V}$ then the training value $V_{train}(b)$ for any intermediate board state $b$ would be $\hat{V}(Successor(b))$.

\begin{equation}
    V_{train}(b) \longleftarrow \hat{V}(Successor(b))
\end{equation}

For board states closer to the end of the game, $\hat{V}$ is more accurate. It is also to note that under specific conditions iterative training value estimation based on successor status estimates can be demonstrated to converge into ideal $V_{train}$ estimates.\cite{TomMMitchell}

\subsection{Weight Adjustment}

After having completed the major components of the design of the generalized algorithm, the last thing that remains is to describe the learning algorithm. This learning algorithm will be used by our agent for adjusting the weights $w_i$($i = 0$ to $i = 6$) to best fit the set of training samples $(b,V_{train}(b))$. The weights will be set such that the squared error(E) is as small as possible. The squared error is calculated between the training values and the values predicted by the hypothesis $\hat{V}$. Hence the error E can be written as - 

$$E = \sum_{(b,V_{train}(b))}^{n} (V_{train} - \hat{V}(b))^2$$

After defining the error that is going to be minimized. It is time to define the algorithm that the agent will use for minimizing $E$ for the observed training examples. There are various algorithms available that could accomplish this purpose, for instance, Frank-Wolfe Method, Hessian-Free Optimization Method, and many which can be applied from \cite{sun2019survey}. For this study $Least Mean Square Training Rule$ is used. This algorithm(LMS) will gradually refine our defined weights with each input training example.
LMS best fits the current estimate of $V$, i.e. $\hat{V}$, for each training example by changing the weights in the direction that reduces the error $E$. \par

"LMS algorithm can also be viewed as performing a stochastic gradient-descent search through the space of possible hypotheses to minimize the squared error $E$"\cite{TomMMitchell}. It can be mathematically written as \par

LMS weight update rule for training example $(b,V_{train}(b))$

\begin{equation}
    w_i \longleftarrow w_i + \eta(V_{train}(b) - \hat{V}(b))x_i
\end{equation}

The learning rate is defined as $\eta$ and is a small constant (e.g 0.3, 0.4, etc). It regulates the scale of weight change. When the error ($V_{train}(b) - \hat{V}(b)$) is zero, no weights are adjusted. It has also been shown in \cite{TomMMitchell} that a standard weight-turning approach converges to the least squared approximation to the  $V_{train}$ values in some settings. \par

After all the components of the tic tac toe agent architecture  defined, it can be summarized as follows:

\begin{itemize}
    \item Task(T): Tic Tac Toe's game play
    \item Performance(P): The number of games won(calculated with win/draw ratio)
    \item Experience(E): Indirect feedback from solution trace
    (game history), generated from playing games against itself
    
    \item Estimating Intermediate training values \par 
    $V_{train}(b) \longleftarrow \hat{V}(Successor(b))$
    \item Final Training Values \par
        $V_{final}(b)$ = 100(win)$|$(draw)$|$-100(loss)
    
    \item LMS Training Rule \par
    $w_i \longleftarrow w_i + \eta(V_{train}(b) - \hat{V}(b))x_i$

\end{itemize}

\section{The Final Design}
After defining all the components of the defined architecture, starting from defining the tic tac toe as a well-posed problem and solving it using \emph{reinforcement learning} as a base and \emph{stochastic-gradient descent} to fine-tune the estimated weight vector parameters for each training example generated during various iterations of several games played against itself. It's time to put it all together. \par

Reinforcement-Learning is like an action-reward system. The agent is rewarded for taking a step in the right direction(In our case, Winning Tic Tac Toe) and penalized for wrong decisions. In our case, the RL Model uses a linear target function but various other possibilities like quadratic, cubic polynomials for a target function also exist. The Figure \ref{RL_Model} below illustrates a typical RL model:

\begin{figure}[h]
    \centering
    \includegraphics[width = 8.8cm]{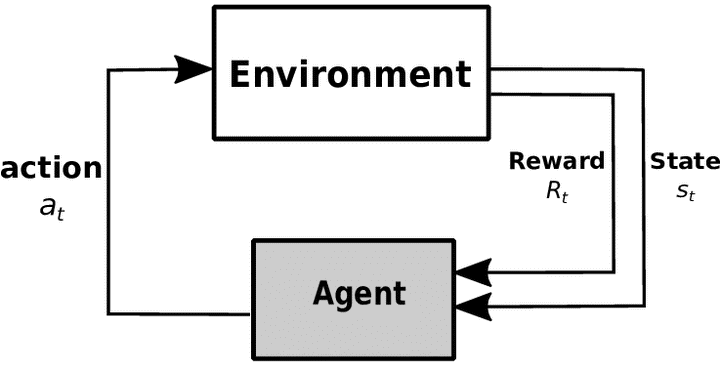}
    \caption{Action-Reward feedback loop for a RL model}
    \label{RL_Model}
    
\end{figure}

Many learning systems in use today use these four key components: Performance System, Critic, Generalizer, and Experiment Generator, and the final design is based on these four components. More information and definitions about these components are given in the following sections.

\begin{itemize}

    \item Performance System - A component that uses the learned objective function to solve the task output task. This module takes a newly generated problem as input, which may be the game's initial state or a randomized state, and generates a game trace of the game played by the agent against itself. Using the current weight vector and an approximation $V_{train}$ of the objective function, it chooses the best move out of all the legal moves available in the current state to produce the game trace.

    \item Critic - A component that corresponds to the training rule given by Equation 2. Takes game trace as an input parameter from Performance system. Generates training examples from the input to be further used by the generalizer. Training examples are generated by appending the feature vector of each state in the game trace with their corresponding utility value.
    
    \item Generalizer -  It corresponds to Least Mean Square Algorithm in Equation (3) and the function is described by Learned weights $w_0...w_6$. As a result of the input training examples, a generalized hypothesis is created that integrates other and existing training examples. Gives an estimate of the target function with $\hat{V}$ as the performance hypothesis.

    \item Experiment Generator - Takes input from Generalizer in the form of a generalized hypothesis and outputs a new problem/state for the Performance system to wander upon. For each iteration, the proposed state may be the same initial state (as in our architecture), a randomized state, or something else entirely. It aims to increase the agent's overall learning rate by providing input samples from different domains, thereby increasing the agent's domain knowledge. Figure \ref{initialState} shows the initial state outputted by the Experiment Generator.    
    
    \begin{figure}[h]
    \centering
    \includegraphics[width = 2cm]{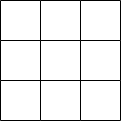}
    \caption{Same initial state returned by the experiment-generator}
    \label{initialState}
    
\end{figure}
\par

The role of each component in the described architecture and its Implementation in an object-oriented manner is described in Figure \ref{architecture} and Figure \ref{classDiagram} respectively.

\begin{figure}[H]
    \centering
    \includegraphics[width = 8.8cm]{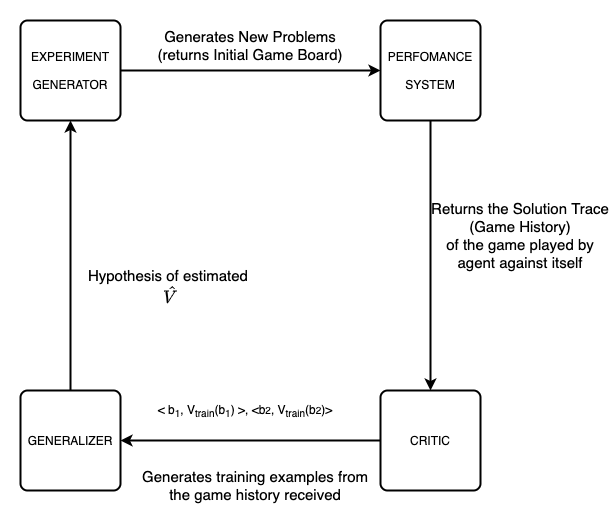}
    \caption{Role of each component in the defined architecture}
    \label{architecture}
    
\end{figure}

\begin{figure}[H]
    \centering
    \includegraphics[width = 8.8cm]{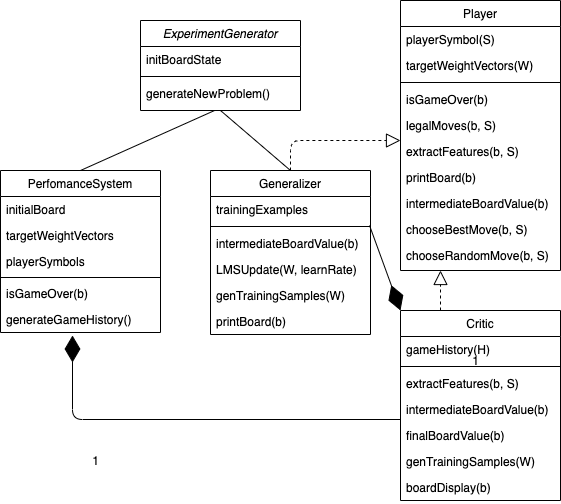}
    \caption{Class Diagram for the defined Architecture}
    \label{classDiagram}
    
\end{figure}

\end{itemize}

\begin{algorithm}[H]
 \caption{Proposed algorithm for Training of Agent}
 \begin{algorithmic}[1]
 \renewcommand{\algorithmicrequire}{\textbf{Input:}}
 \renewcommand{\algorithmicensure}{\textbf{Output:}}
 \REQUIRE numTrainingSamples(n)
 \ENSURE  targetweightVector($W$), numAgentWins($nW$), numAgentLoss($nL$), numAgentDraws($nD$)
 \\ \textit{Initialisation}:
 $W$=[0.5,0.5,0.5,0.5,0.5,0.5,0.5],
 trainGamesCount = 0, \par nW = nL = nD = 0

  \WHILE{trainGamesCount != n}
  \STATE boardStateInitial = expGen.generateNewProblem()

    \STATE Obtain the solution trace of the game played by the agent by choosing \textbf{best(legalmoves(b))}, comparing utility value $\hat{V}(b)$ for each legal move using the current $W$.

    \FOR{board in range(len(gameHistory) - 1)}
    \STATE featureV = extractFeatures(board) //$x_0...x_6$
    
    \STATE calculate the \textbf{approximate utility value} $\hat{V}(b)$ of intermediate board states using featureV and current $W$ using equation (1)    
    \ENDFOR
    
    \STATE calculate the \textbf{utility value} $V(b)$ of the final board state(Loss,Win,Draw)
    
    \STATE  $trainingExamples= [(featureV_1,V_{train}(b_1)),(featureV_2,V_{train}(b_2)),..]$

    \STATE $finalScore = trainingExamples[-1][1]$ 
        \STATE Increment gameStatus counts $nW$, $nL$ and $nD$ based on the Final game result i.e FinalScore
    
    
    
    
    \STATE Update $W$ for all trainingExamples using \textbf{LMS} rule
    
    \FOR{trainExample in trainingExamples}
    \STATE $V_{train}(b) \longleftarrow trainingExample[1]$
    \STATE $featureV \longleftarrow trainingExample[0]$
    \STATE $\hat{V}(b)  \longleftarrow \overrightarrow{W}.featureV$
    \STATE Substitue $V_{train}(b)$ and $\hat{V}(b)$ in eq (3), return the \textbf{new} $W$
    
    \ENDFOR
    
    \STATE $ trainGamesCount \longleftarrow trainGamesCount+1$

  \ENDWHILE
 \RETURN $W, nW, nL, nD$ 
 \end{algorithmic} 
 \end{algorithm}

 \section{Results}

 \begin{table}[ht]

\caption{Agent(X) playing against Itself(O)}
\begin{center}
\begin{tabular}{|c|c|c|c|c|}
\hline
Games Played & Wins & Loss & Draws & Win/Draw\\

\hline
1000 &  761 & 82 & 157 & 4.85 \\
\hline
10000 & 7524 & 827 & 1649 & 4.56 \\
\hline
100000 & 78241 & 8323 & 13436	& 5.82\\
\hline
200000 & 159053	& 16288 & 24659	& 6.45	\\
\hline
300000	& 240136 & 24202 & 35662 & 6.73 \\
\hline
400000 & 326749	& 30979	& 42272 & 7.73\\
\hline
500000 & 410749	& 38336	& 50915 & 8.07 \\
\hline
600000 & 494598	& 45645 & 59757	& 8.28 \\
\hline
700000 & 580906	& 52567 & 66527 & 8.73 \\
\hline
800000 & 667257 &	59554 &	73189 &	9.12 \\
\hline
900000 & 753805 & 66316 & 79879 &9.44 \\
\hline
1000000 & 840239 & 73203 & 86558 & 9.71 \\
\hline
1100000 & 926767 & 79939 & 93294 & 9.93 \\
\hline
1200000	& 1013183 & 86815 & 100002 & 10.13	 \\
\hline
1300000 & 1099240 & 93803 & 106957 & 10.28 \\
\hline
1400000 & 1185745 & 100555 & 113700 & 10.43	 \\
\hline
1500000 & 1272279 & 107315 & 120406 & 10.57 \\
\hline

\end{tabular}
\label{win_draw_table}
\end{center}
\end{table}

\begin{table}[ht]

\caption{Respective Weights vectors for each iteration of Games Played}
\begin{center}
\scalebox{0.9}{
\begin{tabular}{|c|c|c|c|c|c|c|c|}
\hline
Games & w0 & w1 & w2 & w3 & w4 & w5 & w6\\

\hline
1000 & 75.2 &  -8.4 & -166.8 & -39.4 & -171.0 & 55.4 & -115.1 \\
\hline
10000 & 70.6 & -9.9 & -122.3 & 31.0 & -93.2 & 43.0 & -208.4 \\
\hline
100000 & 72.8 & -0.2 &  -23.2 & 54.4 & -158.2 & 45.9 & -208.0\\
\hline
200000 & 44.3 & -9.4 & 1.0	& 38.0 & -85.9 & 56.8 & -175.5	\\
\hline
300000 & 81.4 & -157.9 & -1.1 & 16.9 & -15.4 & 23.6 & -202.8 \\
\hline
400000 &62.2	& -157.9 & -19.0 & 42.8 & -34.0 & 41.0 & -197.1 \\
\hline
500000 & 93.6&	-17.6&	-53.5 & 29.5 & -135.3&	29.8&	-227.4 \\
\hline
600000 & 97.9&	-168.2&	-51.2&	29.0&	-7.2&	51.5&	-218.5 \\
\hline
700000 & 67.3&	-168.2&	-41.9&	27.0&	-47.5&	40.8&	-186.5 \\
\hline
800000 & 77.1&	-168.2&	-2.2&	22.9&	-20.1&	62.5&	-198.9 \\
\hline
900000 & 84.2&	-168.2&	-28.9&	29.4&	10.5&	41.0&	-197.4 \\
\hline
1000000 & 67.0&	-168.2&	11.4&	16.8&	12.0&	34.1&	-197.5 \\
\hline
110000 & 61.7&	-168.2&	-72.9&	21.6&	-20.8&	70.0&	-203.9 \\
\hline
1200000	& 95.4&	-168.2&	-61.6&	76.9&	-25.1&	33.0&	-160.1	 \\
\hline
1300000 & 58.1&	-168.2&	-76.4&	21.6&	-19.4&	36.6&	-183.3 \\
\hline
1400000 & 60.8&	-168.2&	-36.4&	30.6&	-6.0&	47.3&	-209.5 \\
\hline
1500000 & 93.4&	-168.2&	-27.3&	12.4&	-6.0&	18.9&	-198.6 \\
\hline

\end{tabular}}
\label{weight_vector_table}
\end{center}
\end{table}

\begin{figure}[h]
    \centering
    \includegraphics[width = 9cm]{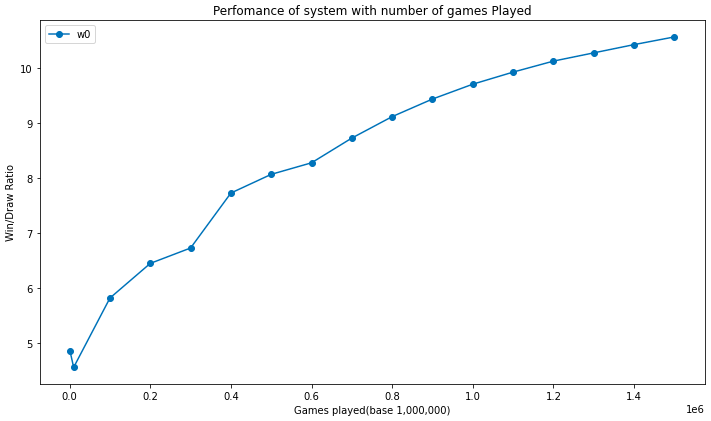}
    \caption{Variation in performance of agent with number of games played(Coloured)}
    \label{win_draw_ratio}
    
\end{figure}

\begin{figure}[h]
    \centering
    \includegraphics[width = 9cm]{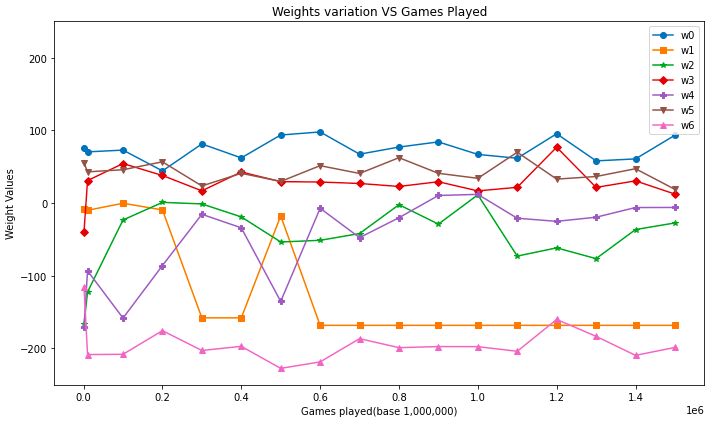}
    \caption{Changes in weights with number of games played(Coloured)}
    \label{weight_graph}
\end{figure}

Tic Tac Toe is a biased game in favor of the first player, giving the player who starts the game more chances to win while leaving the opponent with either a loss or a tie. The game will end in a draw if both players play intelligently. A "Never-Loss" Generalised strategy allows the agent to win or at the very least keep a tie in any game. \par

    


The Learning rate $\eta$ was set to 0.4 on a 3X3 Tic Tac Toe board using Algorithm 1. It was trained on Google Colaboratory for 6 Hours for testing. \par 

Firstly the win ratio was tried after each iteration, which is defined by

\[ \text{Win Ratio}=\frac{\text{Number of games Won}}{\text{Number of Games Played}}  \]

but the results just kept providing ineffable results, so the next thing that was tried was the win/draw ratio, which highlighted the correct output of the agent after each iteration.
\[ \text{Win Draw Ratio}=\frac{\text{Number of games Won}}{\text{Number of Games Drawn}}  \]

Table \ref{win_draw_table} gives us the observation that the agent keeps on improving through the win/draw ratio. Some of the games played by the agent which resulted in Win and draw are given in Appendix in Figures \ref{appendix_draw} and Figure \ref{appendix_win} \par.

Table \ref{weight_vector_table} gives us insight on what features are important to agent it is namely $w5$ which is the Number of Instances of symbol "X" in a row and it is trying to maximize it to whatever extent possible while minimizing $w6$ which results in opponent winning and giving it whatever minimizing number possible. These results are also plotted on graphs in Figure \ref{win_draw_ratio} and \ref{weight_graph}. The most amount of variation is observed in $w1$ in between 0.4 and 0.6(Base 1,000,000) which means the agent trying to explore different starting positions for starting the game. In theory, the first player playing to any corner gives the best winning chances. If the correct play is assumed from both the players then the game will always result in a draw.

 \section{Conclusion and future work}
In this, an attempt was given to propose a generalized Reinforcement Learning algorithm for the game of tic tac toe by defining it as a generalized well-posed problem using gradient descent as stochastic. The algorithm was implemented in 4 main components namely Experiment generator, Performance System, Critic, and Generalizer. 
The proposed algorithm is implemented using an object-oriented paradigm which provides a generalized way of implementing it to larger board states for 4X4, 5X5, etc. \par
The research was started with defining all the features that are to be extracted from a board state, defining feature and weight vectors for each one of them, and finalizing a target function. Defined respective ways to calculate utility values for terminal and intermediate states. Used the LMS weight update rule to update each weight in the weight vector after each training example. The final implementation was tested and an algorithm is proposed. The agent was allowed to play 1,500,000 games against itself with a learning rate of 0.4 and recorded the performance of the agent for intermediate checkpoints. The results of which were quite promising, with an increase of the agent's win/draw ratio. This approach can be easily expanded(Generalized) to higher board states of 4X4, 5X5, etc by defining the respective board features to be extracted. \par

There are a few more conclusions that we can draw from this study:

\begin{itemize}
    \item  The values of weight vectors $w5$ and $w6$ in Table \ref{weight_vector_table} suggest that there is a fair probability of having a no-loss tic tac toe strategy. However, a game like chess, which can also be solved using the state-space search Min-Max algorithm and in which the engines (agents) are constantly changing, cannot be compared to tic tac toe because it is a much more complex game.
    
    \item If the algorithm is extended to other games, changes to the number and types of feature vectors extracted, as well as the Target Function, must be investigated to better match the game. Since the linear target function used in this study may not be the best fit for other complex board games. For those board games Quadratic, Cubic, and other target functions may be needed.

\end{itemize}

In this research, only the training of the agent via Indirect Learning is explored. The future scopes of this study might include Indirect Learning for some portion of learning to complement with direct methods to offer a more comprehensive view of agent learning. But there are still a lot of optimizations that can be done to reduce the training time to reach optimal results via the proposed algorithm. This may inspire other researchers to conduct similar experiments for several other games as this generalized proposed algorithm can also be easily extended to be used to solve other perfect information board games like Checkers, Infinite Chess, etc

\section*{References}

\bibliography{generalized_agent}

\begin{thebibliography}{10}
\expandafter\ifx\csname url\endcsname\relax
  \def\url#1{\texttt{#1}}\fi
\expandafter\ifx\csname urlprefix\endcsname\relax\def\urlprefix{URL }\fi
\expandafter\ifx\csname href\endcsname\relax
  \def\href#1#2{#2} \def\path#1{#1}\fi

\bibitem{games_possible}
How many tic-tac-toe (noughts and crosses) games are possible?,
  \url{http://www.se16.info/hgb/tictactoe.htm}, accessed: 2021-1-15.

\bibitem{chou2013using}
C.-H. Chou, Using tic-tac-toe for learning data mining classifications and
  evaluations, International Journal of Information and Education Technology
  3~(4) (2013) 437.

\bibitem{singh2014never}
A.~Singh, K.~Deep, A.~Nagar, A" never-loose" strategy to play the game of
  tic-tac-toe, in: 2014 International Conference on Soft Computing and Machine
  Intelligence, Vol.~1, IEEE, 2014, pp. 1--5.

\bibitem{maureT}
M.~T. Carroll, S.~T. Dougherty,
  \href{https://doi.org/10.1080/0025570X.2004.11953263}{Tic-tac-toe on a finite
  plane}, Mathematics Magazine 77~(4) (2004) 260--274.
\newblock \href
  {http://arxiv.org/abs/https://doi.org/10.1080/0025570X.2004.11953263}
  {\path{arXiv:https://doi.org/10.1080/0025570X.2004.11953263}}, \href
  {http://dx.doi.org/10.1080/0025570X.2004.11953263}
  {\path{doi:10.1080/0025570X.2004.11953263}}.
\newline\urlprefix\url{https://doi.org/10.1080/0025570X.2004.11953263}

\bibitem{sutton2018reinforcement}
R.~S. Sutton, A.~G. Barto, Reinforcement learning: An introduction, Vol.~1, MIT
  press, 2018.

\bibitem{mnih2015human}
V.~Mnih, K.~Kavukcuoglu, D.~Silver, A.~A. Rusu, J.~Veness, M.~G. Bellemare,
  A.~Graves, M.~Riedmiller, A.~K. Fidjeland, G.~Ostrovski, et~al., Human-level
  control through deep reinforcement learning, nature 518~(7540) (2015)
  529--533.

\bibitem{silver2017mastering}
D.~Silver, T.~Hubert, J.~Schrittwieser, I.~Antonoglou, M.~Lai, A.~Guez,
  M.~Lanctot, L.~Sifre, D.~Kumaran, T.~Graepel, et~al., Mastering chess and
  shogi by self-play with a general reinforcement learning algorithm, arXiv
  preprint arXiv:1712.01815.

\bibitem{qlearning}
D.~H. Widyantoro, Y.~G. Vembrina, Learning to play tic-tac-toe, in: 2009
  International Conference on Electrical Engineering and Informatics, Vol.~1,
  2009, pp. 276--280.
\newblock \href {http://dx.doi.org/10.1109/ICEEI.2009.5254776}
  {\path{doi:10.1109/ICEEI.2009.5254776}}.

\bibitem{wang2021searching}
H.~Wang, Searching by learning: Exploring artificial general intelligence on
  small board games by deep reinforcement learning, Ph.D. thesis, Leiden
  University (2021).

\bibitem{tsividis2021human}
P.~A. Tsividis, J.~Loula, J.~Burga, N.~Foss, A.~Campero, T.~Pouncy, S.~J.
  Gershman, J.~B. Tenenbaum, Human-level reinforcement learning through
  theory-based modeling, exploration, and planning, arXiv preprint
  arXiv:2107.12544.

\bibitem{wang2021adaptive}
H.~Wang, M.~Preuss, A.~Plaat, Adaptive warm-start mcts in alphazero-like deep
  reinforcement learning, in: Pacific Rim International Conference on
  Artificial Intelligence, Vol.~1, Springer, 2021, pp. 60--71.

\bibitem{gu2021enhanced}
B.~Gu, Y.~Sung, Enhanced reinforcement learning method combining one-hot
  encoding-based vectors for cnn-based alternative high-level decisions,
  Applied Sciences 11~(3) (2021) 1291.

\bibitem{scheiermann2022alphazero}
J.~Scheiermann, W.~Konen, Alphazero-inspired general board game learning and
  playing, arXiv preprint arXiv:2204.13307.

\bibitem{bhatt2007002evolution}
A.~Bhatt, P.~Varshney, K.~Deb, Evolution of no-loss strategies for the game of
  tic-tac-toe, IIT, Kanpur, Department of Mechanical Engineering, KanGAL Report
  Number 2007002.

\bibitem{bits_pilani}
N.~F. {Rajani}, G.~{Dar}, R.~{Biswas}, C.~K. {Ramesha}, Solution to the
  tic-tac-toe problem using hamming distance approach in a neural network, in:
  2011 Second International Conference on Intelligent Systems, Modelling and
  Simulation, Vol.~1, 2011, pp. 3--6.
\newblock \href {http://dx.doi.org/10.1109/ISMS.2011.70}
  {\path{doi:10.1109/ISMS.2011.70}}.

\bibitem{fogel1993using}
D.~B. Fogel, Using evolutionary programing to create neural networks that are
  capable of playing tic-tac-toe, in: IEEE International Conference on Neural
  Networks, Vol.~1, IEEE, 1993, pp. 875--880.

\bibitem{karmanova2021swarmplay}
E.~Karmanova, V.~Serpiva, S.~Perminov, A.~Fedoseev, D.~Tsetserukou, Swarmplay:
  Interactive tic-tac-toe board game with swarm of nano-uavs driven by
  reinforcement learning, in: 2021 30th IEEE International Conference on Robot
  \& Human Interactive Communication (RO-MAN), Vol.~1, IEEE, 2021, pp.
  1269--1274.

\bibitem{doi:10.1080/08839514.2021.1934265}
S.~Videgaín, P.~G. Sánchez,
  \href{https://doi.org/10.1080/08839514.2021.1934265}{Performance study of
  minimax and reinforcement learning agents playing the turn-based game iwoki},
  Applied Artificial Intelligence 35~(10) (2021) 717--744.
\newblock \href
  {http://arxiv.org/abs/https://doi.org/10.1080/08839514.2021.1934265}
  {\path{arXiv:https://doi.org/10.1080/08839514.2021.1934265}}, \href
  {http://dx.doi.org/10.1080/08839514.2021.1934265}
  {\path{doi:10.1080/08839514.2021.1934265}}.
\newline\urlprefix\url{https://doi.org/10.1080/08839514.2021.1934265}

\bibitem{acm_borovska}
P.~Borovska, M.~Lazarova, Efficiency of parallel minimax algorithm for game
  tree search, in: Proceedings of the 2007 international conference on Computer
  systems and technologies, Vol.~1, 2007, pp. 1--6.

\bibitem{keswani2022convergent}
V.~Keswani, O.~Mangoubi, S.~Sachdeva, N.~K. Vishnoi, A convergent and
  dimension-independent min-max optimization algorithm, in: International
  Conference on Machine Learning, Vol.~1, PMLR, 2022, pp. 10939--10973.

\bibitem{cai2022accelerated}
Y.~Cai, W.~Zheng, \href{https://arxiv.org/abs/2210.03096}{Accelerated
  single-call methods for constrained min-max optimization}, arXiv preprint
  arXiv:2210.03096\href {http://dx.doi.org/10.48550/ARXIV.2210.03096}
  {\path{doi:10.48550/ARXIV.2210.03096}}.
\newline\urlprefix\url{https://arxiv.org/abs/2210.03096}

\bibitem{abernethy2021last}
J.~Abernethy, K.~A. Lai, A.~Wibisono, Last-iterate convergence rates for
  min-max optimization: Convergence of hamiltonian gradient descent and
  consensus optimization, in: International Conference on Algorithmic Learning
  Theory, Vol.~1, PMLR, 2021, pp. 3--47.

\bibitem{5205101}
S.~{Sriram}, R.~{Vijayarangan}, S.~{Raghuraman}, X.~{Yuan}, Implementing a
  no-loss state in the game of tic-tac-toe using a customized decision tree
  algorithm, in: 2009 International Conference on Information and Automation,
  Vol.~1, 2009, pp. 1211--1216.
\newblock \href {http://dx.doi.org/10.1109/ICINFA.2009.5205101}
  {\path{doi:10.1109/ICINFA.2009.5205101}}.

\bibitem{TomMMitchell}
T.~M. Mitchell, Machine Learning, 1st Edition, McGraw-Hill, Inc., USA, 1997.

\bibitem{goodfellow2016machine}
I.~Goodfellow, Y.~Bengio, A.~Courville, Machine learning basics, Deep learning
  1 (2016) 98--164.

\bibitem{sun2019survey}
S.~Sun, Z.~Cao, H.~Zhu, J.~Zhao, A survey of optimization methods from a
  machine learning perspective, IEEE transactions on cybernetics 50~(8) (2019)
  3668--3681.

\end{thebibliography}

\section{Appendix}

Some of the games played by the agent against itself resulted in a win and draw by X after having played 500,000 games played against itself. This provides us with a graphical overview of what the performance of the agent is like at this iteration. Non Optimal moves are being played by both agents and a win could have been reached sooner(which was corrected further down the iteration of games).

\begin{figure}[h]
    \centering
    \includegraphics[width = 6cm]{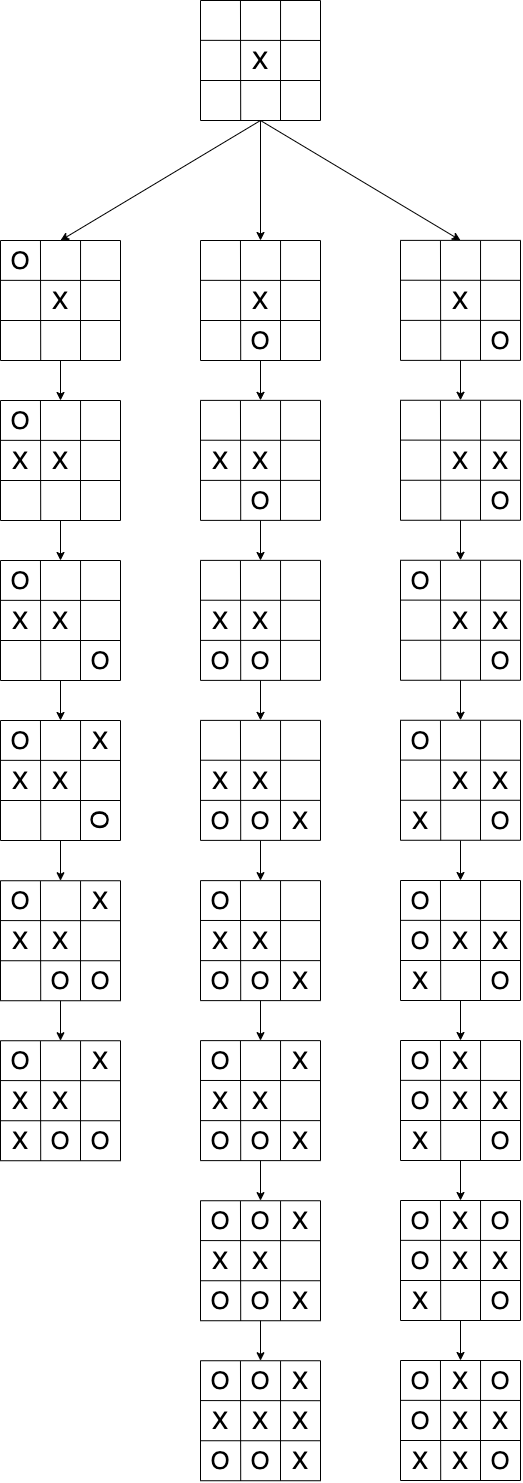}
    \caption{Few games played by the agent that resulted in a win at 500,000 games played}
    \label{appendix_win}
    
\end{figure}

\begin{figure}[h]
    \centering
    \includegraphics[width = 6cm]{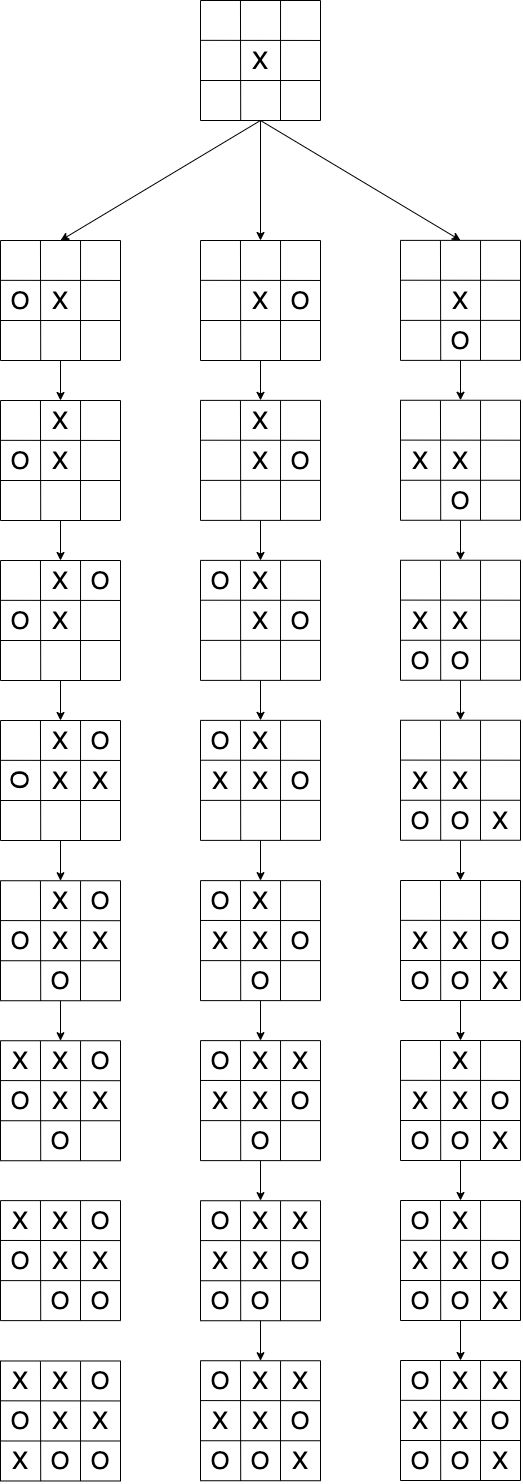}
    \caption{Few games played by the agent that resulted in a draw at 500,000 games played}
    \label{appendix_draw}
    
\end{figure}

\end{document}